\documentclass[a4paper,twocolumn,fleqn]{article}

\usepackage{ist}
\usepackage{graphicx}
\usepackage{amsmath,amssymb}
\usepackage{booktabs}
\usepackage{array}
\usepackage{longtable}
\usepackage{booktabs}
\usepackage{hyperref}
\usepackage{bookmark}

\title{Do Vision Encoders Exhibit Human-like Color Thresholds? }

\author{
Engy Ehab $^{1,2,3}$,
Pablo Hernández-Cámara $^{4}$,
Nahla Belal $^{1}$,
Jesús Malo $^{4}$,
Javier Vazquez-Corral, $^{2,3}$,
Alexandra Gomez-Villa $^{2,3}$\\
$^{1}$ Arab Academy for Science, Technology, and Maritime Transport, Egypt\\
$^{2}$ Computer Vision Center, Spain\\
$^{3}$ Universitat Autònoma de Barcelona, Spain\\
$^{4}$ Image Processing Lab, Universitat de València, Spain
}
\date{}

\begin{document}

\maketitle

\begin{abstract}
Understanding and characterizing human color perception is a longstanding research goal. One of the most traditional approaches is looking for the human color discrimination thresholds, the minimum chromatic differences perceptible to human observers. In recent years, deep neural networks have become the standard networks for computer vision tasks. In particular, deep vision encoders, foundation models trained on large-scale visual data, map images into latent feature representations. Despite the widespread use of deep vision encoders, few studies have investigated whether their internal representations exhibit human-like discrimination thresholds. In this work, we present a large-scale exploratory study probing the chromatic sensitivity of more than 50 pretrained vision encoders, including convolutional networks and vision transformers, against human discrimination thresholds. Using controlled chromatic stimuli at multiple chroma levels, we compare model-derived chromatic discrimination thresholds with human discrimination ellipses through a region-overlap metric (mIoU). Our analysis reveals generally weak alignment between model representations and human perceptual thresholds across all model families, with the best $\text{mIoU} < 0.25$. Moreover, we find that self-supervised encoders consistently outperform supervised ones, while language-supervised models show the most polarized behavior, occupying both the top and bottom of the ranking. These findings suggest that human-like chromatic sensitivity does not emerge naturally from current large-scale visual training objectives for any of the analyzed architectures.
\end{abstract}

\section{Introduction}
\label{sec:intro}

In 1942, MacAdam showed that just-noticeable differences (JNDs), the smallest chromatic differences perceptible to a human observer, are not uniform across the CIE chromaticity diagram. He found that they form ellipses whose size and orientation vary with location \cite{macadam1942visual}. The MacAdam work was lately extended by the BFD dataset~\cite{LuoRigg1986}, 
which combined over 120 discrimination ellipses from 13 independent experiments, and the RIT-DuPont dataset~\cite{Berns1991}, which measured 875 color-difference pairs across 19 CIELAB regions. Both works confirmed and extended these findings to surface colors under industrial viewing conditions. Together, all these works established that the human visual system is perceptually non-uniform: sensitivity to chromatic differences varies across color space, with some regions requiring substantially larger physical changes to produce a perceptible difference than others. 

To address this non-uniformity, the Commission Internationale de l'\'Eclairage (CIE) introduced approximately perceptually uniform color spaces, namely CIELAB and CIELUV, in 1976 \cite{CIE1976}, designed so that Euclidean distances in these spaces would better approximate perceived color differences \cite{Berns2019}. Although they improved perceptual uniformity, Euclidean distances in them still fail to predict discrimination performance consistently across color space \cite{LuoRigg1986}. This limitation motivated progressively refined color-difference formulas, such as CMC \cite{Clarke1984} or CIE94 \cite{CIE1995}, culminating in the CIEDE2000 ($\Delta E_{00}$) \cite{CIEDE2000}, the current CIE standard for perceptual color difference estimation.

While human chromatic discrimination has been extensively characterized through psychophysical experiments, the color representations learned by modern deep vision models remain poorly understood. Contemporary vision encoders, including convolutional neural networks and vision transformers, are trained on massive collections of natural images containing rich and highly structured chromatic statistics. Some works have shown that, as a consequence of training on such large-scale visual data, foundation models develop representations that exhibit  similarities to human visual perception~\cite{gomez2025art, gomez2020color, 10.1167/jov.26.3.14}. However, none of them analyzed the chromatic discrimination thresholds of deep learning models. Yet it remains unclear whether their latent representations reflect the same perceptual non-uniformities observed in human color vision, or whether they organize color according to fundamentally different geometric principles.

In this work, we connect classical color psychophysics with modern computer vision by systematically evaluating the chromatic representation geometry of deep vision encoders against human discrimination behavior. We analyze more than 50 pretrained vision models spanning convolutional architectures, vision transformers, self-supervised encoders, and large-scale multimodal foundation models. Using controlled chromatic stimuli and similarity fields derived from latent feature embeddings, we compare model responses against human discrimination thresholds defined in CIEDE2000 space. This framework allows us to quantify the extent to which modern vision encoders align with human perceptual chromatic thresholds.

\begin{figure*}[!t]
    \centering
    \includegraphics[width=1.0\linewidth]{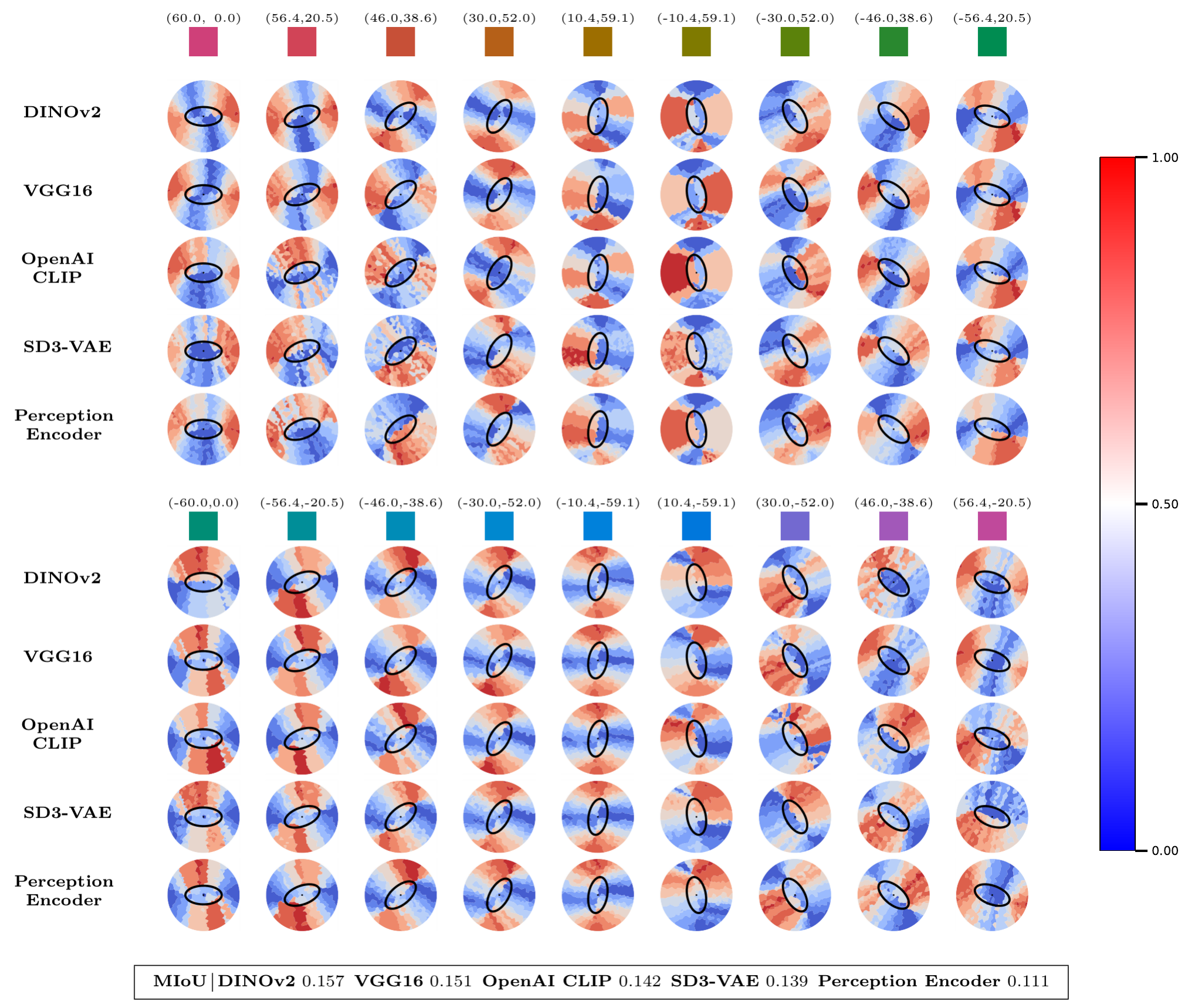}
    \caption{Chromatic similarity fields for representative vision encoders across multiple reference colors at chromaticity 60. The values shown above each reference color chip correspond to its CIELAB (a*, b*). Each column corresponds to a different reference hue, while rows correspond to different model families. Heatmaps visualize cosine distance in the local latent neighborhood around each reference color, with ellipses overlaid to indicate the corresponding human-grounded discrimination regions. Across all architectures, the learned similarity fields exhibit structured and anisotropic organization rather than isotropic chromatic sensitivity. Several models partially recover the psychophysical ellipses, particularly self-supervised encoders such as DINOv2 and Perception Encoder, although substantial mismatches remain in the spatial extent and boundary structure of the similarity regions. The bottom bar reports each model’s mean Intersection over Union (mIoU) between its similarity regions and the human discrimination ellipses, quantifying overall agreement with human chromatic sensitivity, with models ordered left to right by descending mIoU.}
    \label{fig:similarity_fields}
\end{figure*}

\section{Color Representations in Foundation Models}


Recent work has begun to evaluate deep vision models against human psychophysics. In the achromatic domain, Cai et al.~\cite{Cai_2025_CVPR} used feature-space distances to estimate contrast sensitivity in foundation models, a probing strategy similar to the one we adopt here for color. Regarding color specifically, Arias et al.~\cite{arias2025color} showed that Stable Diffusion's latent channels follow a partially disentangled structure, with opponent color axes mirroring the organization observed in human visual perception. Closer to
our work, Wickramanayaka and Oizumi~\cite{wickramanayaka2025systematic} systematically compared the internal color representations of 16 DNNs against human color similarity judgments of 93 colors, finding that only CLIP models achieved fine-grained structural congruence with human perceptual data. Crucially, however, their analysis targets the global representational geometry of color space --- whether models and humans organize the full set of colors
into a similar topological structure. Our work addresses a fundamentally different question: whether models reproduce local chromatic sensitivity, specifically the psychophysically established discrimination thresholds formalized by MacAdam ellipses and CIEDE2000. To our knowledge, no prior work has systematically evaluated this local discrimination geometry in
vision encoders --- this is the gap our work addresses.


\section{Methodology}

\subsection{Vision Encoders}
\label{sec:vis_enc}

We evaluate a large and heterogeneous collection of  50 pretrained vision encoders spanning a broad spectrum of contemporary visual representation learning paradigms. Rather than relying on a single representative per architecture, the benchmark includes multiple variants within each model family, with variability arising from differences in scale, training objectives, and optimization strategies. 
A complete list of the evaluated models and their corresponding training paradigms is provided in Table~\ref{tab:models}. 

The evaluated models include convolutional neural networks (CNNs), vision transformers (ViTs), and hybrid architectures combining convolutional and attention-based components. To analyze the influence of training paradigms on chromatic representation geometry, we include supervised, self-supervised, and multimodal models trained on datasets of varying scale, including ImageNet-1K, ImageNet-12K, and ImageNet-22K. Representative models include OpenCLIP~\cite{radford2021learning, ilharco2021openclip}, DINOv2~\cite{oquab2023dinov2}, Stable Diffusion 3 VAE~\cite{stability2024sd3}, Perception Encoder~\cite{zhang2023peclip}, and VGG16~\cite{simonyan2014vgg}.

\subsection{Stimuli}



\textbf{Reference Colors and Discrimination Regions:} We select 18 reference colors distributed across the chromaticity diagram, whose a*b* values are shown in Figure~\ref{fig:similarity_fields}. Around each reference color, we define a human discrimination region (shown as dashed ellipses in Figure~\ref{fig:similarity_fields}) derived from standard CIE discrimination data, with boundaries determined by the CIEDE2000 metric ($\Delta E_{00}$)~\cite{sharma2005ciede2000}. These regions represent the boundary of perceptual distinguishability and serve as the ground truth against which model representations are evaluated.

To analyze the effect of saturation, stimuli are generated at four chroma levels, $C \in \{5,20,60,80\}$, spanning low- to high-saturation regimes. Luminance is held constant at $L = 50\text{ cd}/\text{m}^{2}$ throughout the experiments to isolate chromatic effects from luminance-driven variations.

\textbf{Stimulus Generation:}
All colors are initially defined in CIELAB space and converted to sRGB~\cite{srgb1996} to ensure compatibility with the input domain of the evaluated models. Each sampled color is rendered into a uniform color patch. All stimuli are resized and normalized according to the preprocessing pipeline associated with each encoder~\cite{radford2021learning, dosovitskiy2020vit}.

\subsection{Evaluation Protocol}

For each reference color, we define a circular sampling region centered on the reference stimulus with a radius equal to twice the semi-major axis of the corresponding CIEDE2000 ellipse. This ensures coverage of both perceptually similar and perceptually distinct colors. Within this region, we uniformly sample 500 chromatic points, with 250 samples located inside the discrimination region and 250 outside.

Each sampled color is rendered as a stimulus and processed by the corresponding vision encoder to obtain a feature embedding. For each reference color, we use the embedding of its central color chip as the reference representation. For example, if architecture \(X\) produces a 512-dimensional embedding, the central chip is encoded as a 512-dimensional vector, and the embedding of each sampled color within the surrounding color space is compared with this reference vector using cosine similarity. The resulting similarity value determines the corresponding point in Fig.\ref{fig:similarity_fields}.
 This procedure produces a  similarity field around each reference color within the encoder's latent representation space. Regions of high similarity correspond to colors treated by the model as perceptually close to the reference, whereas low-similarity regions correspond to colors represented as increasingly distinct. By comparing the structure of these similarity fields with human discrimination thresholds, we can evaluate the extent to which modern vision encoders exhibit human-aligned chromatic thresholds.

\subsection{Evaluation Metrics}

To quantify alignment between model-derived similarity fields and human discrimination thresholds, we compute a region-overlap metric based on intersection over union (IoU, Eq.~\ref{eq:iou}). For each reference color, let $\mathcal{G}$ denote the set of sampled points located inside the human discrimination region. We define a predicted set $\mathcal{P}$ by selecting the 250 sampled points with the lowest cosine distance to the reference embedding from all sampled points, including those both inside and outside the domain delimited by the ground-truth ellipse, ensuring equal cardinality between $\mathcal{G}$ and $\mathcal{P}$ and avoiding size-related biases. The overlap between both sets is measured as:

\begin{equation}
\hspace{2cm}\text{IoU} = \frac{|\mathcal{G} \cap \mathcal{P}|}{|\mathcal{G} \cup 
\mathcal{P}|}
\label{eq:iou}
\end{equation}

We report the mean IoU (mIoU) across all reference colors and chroma levels as the primary measure of perceptual alignment, where 1 indicates perfect overlap between model and human discrimination regions and 0 indicates no intersection. This metric is used to rank all evaluated encoders and to identify the top-performing models for subsequent qualitative and comparative analyses.

\section{Results}
\label{sec:exp}

\subsection{Structure of Model-Derived Similarity Fields}

First, to better understand the lately analyzed mIoU scores, Figure~\ref{fig:similarity_fields} visualizes chromatic similarity fields for some representative encoders across multiple reference colors. Each map shows the organization of the encoder latent space around a reference hue, while dashed ellipses indicate the corresponding human discrimination regions.

A central observation is that all evaluated encoders exhibit highly structured and strongly anisotropic chromatic similarity fields. Rather than organizing colors isotropically around each reference point, the latent spaces consistently develop preferred directions of chromatic similarity whose sizes and orientations vary across hue. This demonstrates that modern vision encoders do not treat color uniformly in latent space, but instead learn structured chromatic organizations.

Notably, the structures of the similarity fields often partially align with the orientation of the human discrimination ellipses. This effect is particularly visible for self-supervised models such as DINOv2 and the Perception Encoder, whose latent similarity structures frequently follow the principal axes of the perceptual ellipses. However, despite this partial directional agreement, the spatial extent and boundary structure of the learned similarity regions typically diverge from the psychophysical human regions. In many cases, the discrimination ellipse intersects regions of rapidly changing similarity rather than lying within a coherent high-similarity region.

The similarity fields also reveal substantial differences between representation-learning paradigms. OpenCLIP models produce comparatively smooth and spatially diffuse fields, consistent with a semantically compressed latent organization in which subtle chromatic distinctions are deemphasized. In contrast, DINOv2 and Perception Encoder representations exhibit sharper and more locally adaptive chromatic structure, suggesting that self-supervised objectives preserve low-level chromatic relationships more effectively. Representations derived from Stable Diffusion VAEs appear noisier and less geometrically stable, with fragmented similarity regions and weaker agreement with the psychophysical ellipses.

A further consistent pattern is the strong dependence on chromatic location: the geometry of the similarity fields changes substantially across hue, qualitatively mirroring the location-dependent non-uniformity of human perceptual discrimination. However, despite this structural resemblance, the precise geometric organization of the learned fields remains substantially misaligned with the human references, suggesting that models capture the existence of chromatic non-uniformity without accurately reproducing its spatial structure.

Taken together, these visualizations suggest that deep vision encoders do not fail to develop chromatic structure altogether. Instead, they consistently learn anisotropic chromatic geometries, but they only partially overlap with human perceptual regions.

\begin{figure*}[!t]
    \centering
    \includegraphics[width=\linewidth]{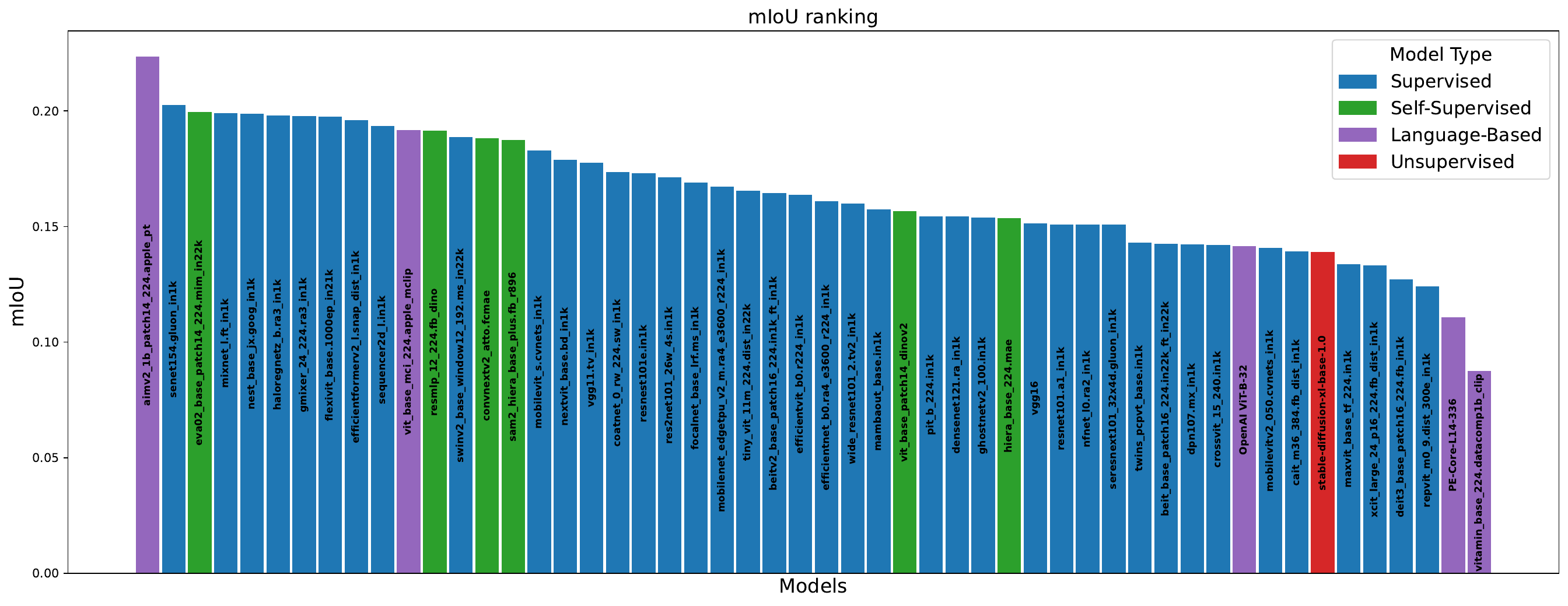}
    \caption{Mean intersection over union (mIoU) between model and human chromatic discrimination thresholds for all evaluated vision encoders. Models are colored according to training paradigm: supervised, self-supervised, language-supervised, and unsupervised (VAE-based) goals. Higher mIoU indicates stronger alignment between the model and human discrimination thresholds.}
    \label{fig:model_ranking}
\end{figure*}

\subsection{Overall Human Chromatic Discrimination Alignment}

We then evaluate the global alignment between model-derived and human discrimination thresholds using the mean intersection over union (mIoU). Figure~\ref{fig:model_ranking} summarizes the performance of all evaluated encoders.

Across the entire benchmark, alignment with human perceptual discrimination remains relatively limited. Even the highest-performing models achieve only modest overlap with the human discrimination regions ($\text{mIoU} \approx 0.22$), indicating that current visual representation learning objectives and architectures do not naturally produce human-like chromatic thresholds. Nevertheless, the results reveal trends across training paradigms.

Self-supervised encoders generally achieve the strongest alignment scores, suggesting that objectives preserving local image structure and visual consistency may better retain more perceptually human-like chromatic thresholds. Language-supervised models show the most polarized behavior, occupying both the top and bottom of the ranking: AIM-v2 achieves the highest alignment across all evaluated encoders (mIoU$ \approx 0.22$ ), while several other language-supervised models rank among the lowest. The unsupervised model (Stable Diffusion VAE) exhibits weak alignment with human discrimination thresholds. Supervised models show the highest variability, indicating that the training objective alone does not fully explain alignment with human thresholds.

Interestingly, several classical supervised convolutional architectures, such as VGG16, remain competitive with substantially larger and more recent transformer-based foundation models. This observation indicates that increases in model scale and representational capacity do not necessarily translate into improved perceptual chromatic organization.

Finally, it is interesting to highlight that the overall distribution of scores is relatively low. While differences arises between model training goals, no encoder approaches near-human agreement with the human discrimination thresholds. This inconsistency suggests that the observed limitations may reflect broader properties of contemporary large-scale visual training objectives and architectural deficiencies.

\subsection{Dependence on Hue Region and Chroma Level}

\begin{figure*}[!t]
\centering
    \includegraphics[width=\linewidth]{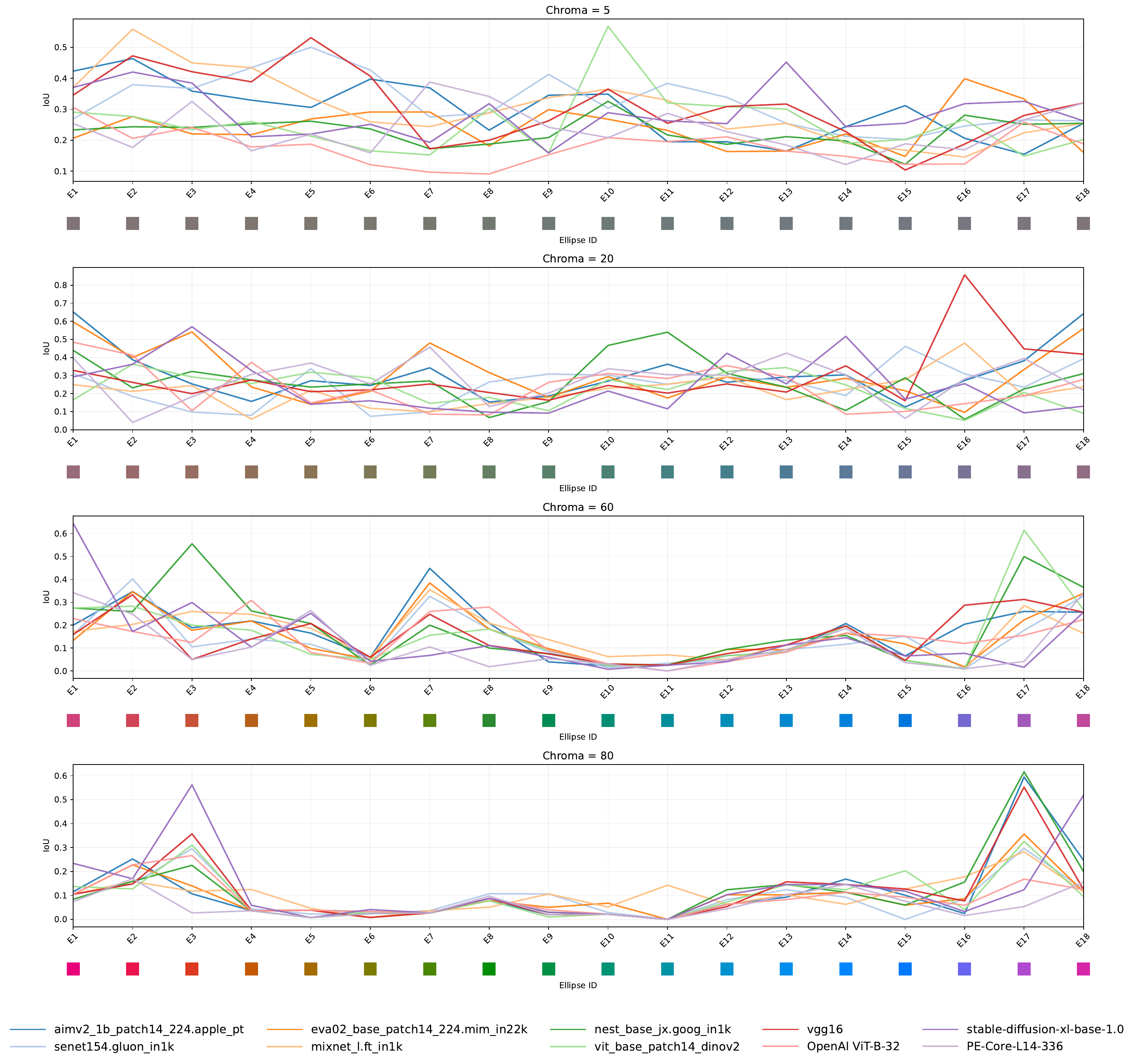}
    \caption{Intersection over union (IoU) scores across hue and chroma levels for representative vision encoders. Each subplot corresponds to a different chroma level, while curves show IoU variation across hue ellipses. The results reveal strong dependence on both chromatic location and saturation regime.}
    \label{fig:per_ellipse_iou}
\end{figure*}

Finally, to analyze the dependence on chroma, we check how the perceptual alignment changes across chromatic conditions. Figure~\ref{fig:per_ellipse_iou} reports IoU scores for representative models across hue and chroma levels.

The results reveal that alignment is highly dependent on both chromatic location and saturation regime. Certain hue regions (particularly blues, which are also known to exhibit strong perceptual non-uniformities in classical psychophysical datasets) consistently produce lower IoU scores across nearly all encoders, indicating that these areas of color space are systematically more difficult for learned representations to organize in a psychophysically consistent manner. This behavior is especially pronounced at high chroma levels, where the human discrimination ellipses are larger and their anisotropic structure more complex. Conversely, at lower chroma levels the discrimination regions are smaller, which geometrically facilitates higher IoU scores regardless of the model's actual chromatic sensitivity.

Despite these local variations, the relative ranking between models remains remarkably stable across hue regions and chroma levels. Models that achieve strong global mIoU scores also maintain comparatively robust performance across most discrimination regions. Conversely, lower-performing architectures exhibit larger fluctuations and more frequent region-specific failures. This consistency indicates that the global ranking is not driven by isolated chromatic regions but instead reflects stable structural properties of the learned latent geometries.

The results further demonstrate a strong dependence on chroma. Alignment patterns vary substantially between low- and high-saturation conditions, revealing that chromatic representation geometry changes non-uniformly across saturation regimes. High chroma levels generally produce more stable alignment, whereas low-chroma conditions exhibit larger variability and more pronounced model failures.

\section{Discussion}
\label{sec:discussion}

The results of this work demonstrate a profound and systematic divergence between modern deep vision encoders and the human chromatic thresholds. While all evaluated models develop anisotropic, structured similarity fields, confirming that they do not treat color space uniformly, these learned thresholds fail to align robustly with human ones. Even the top-performing architectures achieve modest alignment ($\text{mIoU} < 0.25$), revealing a critical gap in contemporary representation learning.

A particularly striking finding is that scaling model size or learning capacity does not naturally yield increase the alignment with human chromatic sensitivity. Classical convolutional networks, such as VGG16, remain highly competitive with significantly larger, modern transformer-based foundation models. This suggests that the emergence of human-like chromatic discrimination thresholds is not governed by architecture type or parameter count. However, the superiority of self-supervised models over purely supervised shows a clear dependence and the importance of the optimization goal.

These findings possess important implications for the deployment of foundation models in human-centric applications. For tasks requiring precise color matching, digital art preservation, or fine-grained product inspection, relying solely on standard frozen embeddings may introduce systemic errors, particularly in highly saturated or non-uniform regions of the color space. If a model treats a perceptually distinct color change as a region of high latent similarity, downstream tasks will inherently fail to preserve human-aligned color fidelity.

\section{Conclusions}

In this work, we presented a large-scale study of chromatic representation geometry in modern vision encoders, evaluating 50 pretrained models against human chromatic discrimination regions. Using controlled chromatic stimuli and similarity-based analysis in latent feature spaces, we quantified the degree to which learned representations align with human perceptual discrimination structure.

Our results show that while deep vision encoders consistently develop structured and anisotropic chromatic similarity geometries, these representations only partially align with human perceptual models. Alignment varies systematically across chromaticity regions and saturation levels, and depends strongly on training paradigm, with self-supervised and supervised models outperforming language-supervised and generative approaches. However, even the best-performing models remain far from human-level chromatic discrimination, suggesting that human-like chromatic discrimination does not naturally emerge from current large-scale visual training objectives.


\textbf{Limitations and future work:} This study has some limitations. First, it focuses on static, synthetic stimuli designed to isolate chromatic effects under controlled conditions. While this enables precise comparison with MacAdam human discrimination thresholds, it necessarily abstracts away from richer visual contexts present in natural scenes. Extending the evaluation to include structured stimuli, such as textured surfaces and object-based renderings, would help determine whether chromatic alignment improves under more ecologically valid conditions. In particular, a promising direction is to quantify whether models preserve the directional structure of MacAdam-like ellipses across different spatial contexts, including full objects rather than flat color patches. Such an analysis would help disentangle whether current models encode chromatic sensitivity primarily at the pixel level or whether these properties persist in higher-level representations.

Additionally, our evaluation is based on cosine distance in frozen feature spaces, which may not fully capture nonlinear decoding processes in downstream tasks. future work could directly compare the \emph{orientation} and \emph{eccentricity} of model-induced similarity fields with psychophysical discrimination ellipses, providing a more fine-grained characterization of geometric alignment.

\section*{Acknowledgments}

The work was partially funded by Grants PID2023-152133NB-I00, PID2024-162555OB-I00; by MICIU/AEI/10.13039/501100011033, ERDF/EU and the FEDER; by Spanish MIU under FPU21/02256; by the grant Càtedra ENIA UAB-Cruïlla (TSI-100929-2023-
2) from the Ministry of Economic Affairs and Digital Transition of Spain; by the grant BBVA Foundations of Science program: Maths, Stats, Comp. Sci. and AI (VIS4NN). JVC also acknowledges the 2025 Leonardo Grant for Scientific Research and Cultural Creation from the BBVA Foundation. The BBVA Foundation accepts no responsibility for the opinions, statements and contents included in the project and/or the results thereof, which are entirely the responsibility of the authors. We acknowledge the EuroHPC Joint Undertaking for awarding us access to Leonardo at CINECA, Italy, under project EHPC-DEV-2026D02-234, and the RES resources provided by BSC on MareNostrum5 under project IM-2026-1-0062.

{\small
\bibliographystyle{unsrt}
\bibliography{ref}
}


\newpage
\appendix
\section{Technical appendices and supplementary material}
\label{appendice}

\begin{table}[h]
\caption{Evaluated models and their training paradigms.}
\vspace{20pt}
\label{tab:models}
\centering
\resizebox{0.5\textwidth}{!}{
\begin{tabular}{ll}
\toprule
\textbf{Model} & \textbf{Training Paradigm} \\
\midrule
senet154.gluon\_in1k          & Supervised \\
mixnet\_l.ft\_in1k            & Supervised \\
nest\_base\_jx.goog\_in1k     & Supervised \\
haloregnetz\_b.ra3\_in1k      & Supervised \\
gmixer\_24\_224.ra3\_in1k     & Supervised \\
efficientformerv2\_l.snap\_dist\_in1k & Supervised \\
sequencer2d\_l.in1k           & Supervised \\
flexivit\_base.1000ep\_in21k  & Supervised \\
swinv2\_base\_window12\_192.ms\_in22k & Supervised \\
beitv2\_base\_patch16\_224.in1k\_ft\_in1k & Supervised \\
beit\_base\_patch16\_224.in22k\_ft\_in22k & Supervised \\
deit3\_base\_patch16\_224.fb\_in1k & Supervised \\
twins\_pcpvt\_base.in1k       & Supervised \\
xcit\_large\_24\_p16\_224.fb\_dist\_in1k & Supervised \\
cait\_m36\_384.fb\_dist\_in1k & Supervised \\
crossvit\_15\_240.in1k        & Supervised \\
pit\_b\_224.in1k              & Supervised \\
vgg16                         & Supervised \\
vgg11.tv\_in1k                & Supervised \\
densenet121.ra\_in1k          & Supervised \\
resnet101.a1\_in1k            & Supervised \\
wide\_resnet101\_2.tv2\_in1k  & Supervised \\
res2net101\_26w\_4s.in1k      & Supervised \\
seresnext101\_32x4d.gluon\_in1k & Supervised \\
resnest101e.in1k              & Supervised \\
nfnet\_l0.ra2\_in1k           & Supervised \\
dpn107.mx\_in1k               & Supervised \\
mobilevit\_s.cvnets\_in1k     & Supervised \\
mobilevitv2\_050.cvnets\_in1k & Supervised \\
ghostnetv2\_100.in1k          & Supervised \\
efficientnet\_b0.ra4\_e3600\_r224\_in1k & Supervised \\
efficientvit\_b0.r224\_in1k   & Supervised \\
tiny\_vit\_11m\_224.dist\_in22k & Supervised \\
mobilenet\_edgetpu\_v2\_m.ra4\_e3600\_r224\_in1k & Supervised \\
repvit\_m0\_9.dist\_300e\_in1k & Supervised \\
coatnet\_0\_rw\_224.sw\_in1k  & Supervised \\
nextvit\_base.bd\_in1k        & Supervised \\
mambaout\_base.in1k           & Supervised \\
maxvit\_base\_tf\_224.in1k    & Supervised \\
focalnet\_base\_lrf.ms\_in1k  & Supervised \\
resmlp\_12\_224.fb\_dino      & Self-Supervised \\
vit\_base\_patch14\_dinov2    & Self-Supervised \\
convnextv2\_atto.fcmae        & Self-Supervised \\
sam2\_hiera\_base\_plus.fb\_r896 & Self-Supervised \\
eva02\_base\_patch14\_224.mim\_in22k & Self-Supervised \\
hiera\_base\_224.mae          & Self-Supervised \\
OpenAI ViT-B-32               & Language-Based \\
vit\_base\_mci\_224.apple\_mclip & Language-Based \\
vitamin\_base\_224.datacomp1b\_clip & Language-Based \\
aimv2\_1b\_patch14\_224.apple\_pt & Language-Based \\
PE-Core-L14-336               & Language-Based \\
stable-diffusion-xl-base-1.0  & Unsupervised \\
\bottomrule
\end{tabular}%
}
\end{table}
\end{document}